\documentclass[journal,twoside,web]
{ieeecolor}
\usepackage{generic}          % JBHI style file (ships with the JBHI template ZIP)
\usepackage{cite}
\usepackage{amsmath,amssymb,amsfonts}
\usepackage{algorithm,algorithmic}
\usepackage{graphicx}
\usepackage{textcomp}
\usepackage{threeparttable}
\usepackage{pifont}
\usepackage{multirow}
\usepackage{booktabs}
\usepackage{hyperref}
\hypersetup{hidelinks=true}
\usepackage{comment}

\newcommand{\cmark}{\ding{51}}
\newcommand{\xmark}{\ding{55}}

\def\BibTeX{{\rm B\kern-.05em{\sc i\kern-.025em b}\kern-.08em
    T\kern-.1667em\lower.7ex\hbox{E}\kern-.125emX}}

\begin{document}

% ---------------------------------------------------------------------
%  TITLE
% ---------------------------------------------------------------------
\title{An Ultra-Widefield Swept-Source OCTA Dataset and a Polar-Gated Mamba Network for Retinal Vessel Segmentation}

% ---------------------------------------------------------------------
%  AUTHORS / AFFILIATIONS / FUNDING
% ---------------------------------------------------------------------
\author{Yang Liu, Yibing Shen, Keming Zhao, Cenk Jiang, Zhenghang Qian,
        Zhicheng Du, Chen~Xiong, Qidong~Shao, Zijun~Lin, Yunqi~Hu,
        Jingjing~Zhou, Lian~Zhang, Peter~E.~Lobie, Peiwu~Qin, and
        \mbox{Chengming Yang}\hfill\mbox{}%
\thanks{This work was supported in part by the Joint TCM Science \&
        Technology Projects of National Demonstration Zones for
        Comprehensive TCM Reform, GZY-KJS-ZJ-2026-032; National Natural
        Science Foundation of China (No.22306048, 62473005, U24A20761);
        Joint Fund of Zhejiang Provincial Natural Science Foundation of
        China, LLSQN26H090003; Sichuan Provincial Natural Science
        Foundation of China, 2026NSFSC1501; Medical and Health Science
        Program of Zhejiang Province, 2025HY1542; Zhejiang Province
        Post-Doctoral Research Project Selected Funding, Z12024150;
        Shenzhen Science and Technology Program (No.JCYJ20250604145715020,
        JCYJ20230807113017035, JCYJ20240813112016022); and Shenzhen
        High-Level and Urgently Needed Talent Startup Funding.
        \textit{(Yang Liu and Yibing Shen contributed equally to this work.)}
        \textit{(Corresponding authors: Peiwu Qin; Chengming Yang)}}%
\thanks{Yang Liu, Yibing Shen, Keming Zhao, Cenk Jiang, Zhenghang Qian,
        Zhicheng Du, Chen Xiong, Qidong Shao, Zijun Lin, Yunqi Hu,
        and Peter E. Lobie are with Tsinghua Shenzhen International
        Graduate School, Shenzhen 518109, China
        (e-mail: lyang22@mails.tsinghua.edu.cn;
        shenyb25@mails.tsinghua.edu.cn;
        rd1tm1jb9um7pt5@outlook.com;
        cenkjiang10@mails.tsinghua.edu.cn;
        qianzx21@mails.tsinghua.edu.cn;
        Wall\_Dalton4xmtrrsse@gmx.co.uk;
        llzzmm0218@163.com;
        wmda130411@gmail.com;
        Xi0218@hotmail.com;
        yunqihu26@126.com;
        pelobie@sz.tsinghua.edu.cn).}%
\thanks{Peiwu Qin is with the Chinese Medicine Guangdong
        Laboratory/Hengqin Laboratory, Hengqin, Guangdong 519031, China
        (e-mail: pwqin1979@gmail.com).}%
\thanks{Jingjing Zhou is with Tsinghua University, Beijing, China
        (e-mail: 18801256889@163.com).}%
\thanks{Lian Zhang is with the Medical Artificial Intelligence Lab,
        The First Hospital of Hebei Medical University, Hebei Medical
        University, Shijiazhuang 050000, China
        (e-mail: lianzhang@hebmu.edu.cn).}%
\thanks{Chengming Yang is with Southern University of Science and
        Technology Hospital, Shenzhen 518055, China
        (e-mail: xluckyyscholar@163.com).}%
}

\maketitle

% ---------------------------------------------------------------------
%  ABSTRACT
% ---------------------------------------------------------------------
\begin{abstract}
Ultra-widefield (UWF) swept-source optical coherence tomography angiography (SS-OCTA) enables large-area retinal vascular imaging, yet vessel segmentation at this scale lacks dedicated public benchmarks and comprehensive evaluation for quantitative vascular analysis. We introduce WOIVES, to our knowledge the first publicly available UWF SS-OCTA vessel-segmentation dataset, comprising 206 eyes from 152 participants with a 24$\times$20~mm$^2$ field of view. WOIVES spans emmetropia to high myopia and provides soft probability vessel annotations. We further propose PG-Mamba, a visual state space model that enhances conventional directional scans with two complementary polar-coordinate scan orders. An auxiliary Dynamic FOV Gating module performs spatial modulation at the bottleneck. PG-Mamba outperformed seven competitive approaches on broad segmentation metrics under cross-validation. It achieved the lowest median absolute errors for vessel density, fractal dimension, and vessel length density. WOIVES is publicly available on Zenodo (DOI: 10.5281/zenodo.21904672), and the PG-Mamba code is available at \url{https://github.com/syb1234567/PG-Mamba}.
\end{abstract}


% =====================================================================
\section{Introduction}
\label{sec:introduction}

\IEEEPARstart{M}{yopia}, particularly high myopia, is a major cause of irreversible visual impairment worldwide, with nearly half of the global population projected to be myopic and approximately one in ten highly myopic by 2050~\cite{holden2016global}. High myopia is characterized by axial elongation accompanied by structural and vascular alterations in the retina and choroid~\cite{baird2020myopia,neelam2012choroidal}. Because these changes are not confined to the posterior pole, assessment over a wider retinal field is important for characterizing myopia-related vascular abnormalities~\cite{he2019association,flores2013relationship,liu2025uwfmyopia}. Optical coherence tomography angiography (OCTA) provides non-invasive, depth-resolved visualization of the retinal and choroidal vasculature and has become an important modality for quantitative vascular analysis~\cite{invernizzi2020imaging}.

Ultra-widefield (UWF) swept-source OCTA (SS-OCTA) further extends this capability by providing large-area, depth-resolved angiography in a single acquisition. With fields of view reaching 24$\times$20~mm$^2$ ($\sim$120\textdegree), UWF SS-OCTA enables simultaneous assessment of central and peripheral vascular patterns~\cite{sampson2022towards,zheng2023advances}. Quantitative vascular measurements derived from these images, including vessel density (VD), vessel length density (VLD), fractal dimension (FD), and connectivity-related measures, depend on accurate vessel segmentation~\cite{li2017retinal,yao2020quantitative}. Segmentation errors in thin vessels and uncertain boundary regions can therefore affect downstream vascular measurements.

Despite the increasing availability of widefield OCTA imaging, current public segmentation benchmarks remain dominated by conventional fields of view. OCTA-500~\cite{li2020octa500}, ROSE~\cite{ma2021rose}, and DRAC~\cite{shang2022drac}, for example, cover fields of view no larger than 12$\times$12~mm$^2$ and do not provide the combination of UWF SS-OCTA vessel annotations and ocular biometric information needed for myopia-oriented analysis. Existing segmentation approaches, including convolutional networks~\cite{ronneberger2015unet,zhou2018unetpp}, Transformer-based models~\cite{cao2022swinunet}, and visual state space models~\cite{gu2023mamba,ruan2024vmunet}, have also been developed and evaluated primarily on conventional imaging fields. For visual state space models in particular, two-dimensional features must be serialized into one-dimensional sequences before selective state-space processing, making the scan order an important architectural choice. Conventional directional scans provide a limited set of predefined spatial orderings, motivating the exploration of complementary scan patterns for UWF vascular segmentation. In addition, segmentation performance is commonly summarized using binary overlap metrics, while continuous annotation agreement, vascular structural consistency, and the relationship between segmentation performance and downstream vascular measurements remain less well characterized.

The main contributions of this work are threefold:
\begin{enumerate}
    \item We introduce WOIVES, to our knowledge the first publicly available UWF SS-OCTA vessel-segmentation dataset, comprising 206 eyes from 152 participants with 480~mm$^2$ coverage and soft probability vessel annotations generated through multi-annotator fusion and expert refinement.
    
    \item We propose PG-Mamba, a U-shaped visual state space model that augments conventional directional scans with two complementary polar-coordinate scan orders. A Dynamic FOV Gating mechanism is introduced for adaptive spatial feature modulation at the bottleneck.
    
    \item Through subject-level five-fold cross-validation and per-eye paired analysis, we benchmark PG-Mamba against seven competitive baselines using overlap, topology-aware, and continuous soft-label metrics, and further assess downstream vascular-measurement agreement. PG-Mamba elevates segmentation performance across five of the six primary metrics and achieves the lowest annotation-derived errors for VD, FD, and VLD.
\end{enumerate}

% =====================================================================
\section{Related Work}
\label{sec:related_work}

\subsection{OCTA Datasets}
\label{sec:rw_datasets}

Public OCTA datasets have played an important role in automated retinal vascular analysis. OCTA-500~\cite{li2020octa500} provides OCTA images acquired at 3$\times$3 and 6$\times$6~mm$^2$ fields of view (FOVs), together with annotations for retinal vascular and disease-related analysis. ROSE~\cite{ma2021rose} contains two OCTA subsets acquired at 3$\times$3~mm$^2$ and provides vessel annotations for retinal vascular segmentation. DRAC~\cite{shang2022drac} extends the imaging area to 12$\times$12~mm$^2$ and supports diabetic-retinopathy-related lesion segmentation, image quality assessment, and grading. Although substantially wider than conventional macular OCTA scans, its 144~mm$^2$ imaging area remains considerably smaller than the 24$\times$20~mm$^2$ acquisitions considered in this study, and its segmentation annotations primarily target pathological lesions rather than the complete vascular tree.

These datasets provide valuable benchmarks for OCTA analysis but do not offer a public UWF SS-OCTA vessel-segmentation resource that combines large-field vascular annotations with ocular biometric information for myopia-oriented analysis. Existing vessel annotations are also predominantly represented as binary masks, which do not retain inter-annotator variation at uncertain vessel boundaries. WOIVES complements these resources by providing 480~mm$^2$ UWF SS-OCTA acquisitions, vessel annotations retained as soft probability maps, and ocular biometry for a clinically characterized subset.

\subsection{Retinal Vessel Segmentation}
\label{sec:rw_seg}

U-Net~\cite{ronneberger2015unet} established the encoder--decoder architecture with skip connections as a standard framework for medical image segmentation. Subsequent CNN-based variants introduced nested skip connections (UNet++~\cite{zhou2018unetpp}), recurrent residual learning (R2U-Net~\cite{alom2018r2unet}), and attention mechanisms (Attention U-Net~\cite{oktay2018attention}). OCTA-specific methods include OCTA-Net~\cite{ma2021rose}, which progressively refines vessel predictions, IPN-V2~\cite{li2020octa500}, which jointly exploits volumetric OCTA information for en face analysis, and OCT2Former~\cite{tan2023oct2former}, which incorporates Transformer-based representations for OCTA segmentation.

Transformer architectures such as Swin-UNet~\cite{cao2022swinunet} and H2Former~\cite{he2023h2former} further improve contextual modeling in medical image segmentation, although maintaining efficient feature modeling becomes increasingly important as spatial resolution and imaging area increase. In parallel, topology-aware objectives such as clDice~\cite{shit2021cldice} have been proposed to evaluate and preserve the connectivity of tubular structures. Most of these approaches, however, have been developed and evaluated on conventional imaging fields and do not explicitly address the combination of large-field OCTA segmentation, soft vascular annotations, and downstream vascular quantification considered in this work.

\subsection{State Space Models in Medical Imaging}
\label{sec:rw_ssm}

Mamba~\cite{gu2023mamba} introduced input-dependent state space modeling for efficient long-sequence processing. VMamba~\cite{liu2024vmamba} extended this formulation to vision through the Visual State Space block and two-dimensional selective scanning, while VM-UNet~\cite{ruan2024vmunet} incorporated visual state space blocks into a U-shaped medical image segmentation framework. Subsequent studies have explored hybrid CNN--Mamba architectures, pure Mamba segmentation networks, large-kernel designs, and adaptive feature modeling~\cite{ma2024umamba,wang2024mambaunet,wang2024lkmunet,yang2024acmamba}. SegMamba further investigated multi-orientation state-space modeling for medical image segmentation~\cite{xing2024segmamba}.

An emerging line of work has shown that the ordering used to serialize two-dimensional features is itself an important design choice. LocalMamba~\cite{huang2024localmamba}, for example, introduces windowed selective scanning and searches scan configurations across network layers to better preserve spatial dependencies. These developments suggest that scan design can influence how spatial relationships are presented to the selective state-space operator. PG-Mamba follows this direction but focuses on UWF OCTA vessel segmentation, augmenting horizontal, vertical, and diagonal scans with two polar-coordinate-based scan orders. Dynamic FOV Gating is additionally used as an auxiliary bottleneck modulation module.

\subsection{Soft Labels and Probability-Level Evaluation}
\label{sec:rw_uncertainty}

Medical image segmentation is commonly trained and evaluated using binary annotations, although uncertain boundaries and inter-annotator variability can result in multiple plausible delineations. Soft-label approaches retain continuous annotation information rather than collapsing it into a single hard target. SoftSeg~\cite{gros2021softseg} demonstrated the utility of soft targets for medical image segmentation, while Louren\c{c}o-Silva and Oliveira~\cite{lourenco2022softlabels} constructed soft labels from multiple expert annotations to represent segmentation variability.

Prediction quality is also not fully characterized by thresholded overlap measures alone. Previous work has examined confidence calibration and predictive uncertainty in medical segmentation~\cite{guo2017calibration,mehrtash2020confidence}, while probability-sensitive scoring rules such as the Brier score~\cite{brier1950verification} provide complementary information about continuous prediction errors. In OCTA vessel segmentation, evaluation has predominantly emphasized overlap-based performance, with less attention to the relationship among soft annotation agreement, probability-level error, vessel caliber, topology, and downstream vascular measurements. We therefore complement binary Dice and IoU with soft Dice, MAE, Brier score, clDice, caliber-stratified analysis, and errors in annotation-derived vascular measurements.

% =====================================================================
\section{WOIVES Dataset}
\label{sec:dataset}

This section describes the construction of WOIVES, a \emph{Widefield swept-source OCTA Image dataset based on VEssel Segmentation}, including data acquisition, image processing, the annotation pipeline, and dataset statistics.

\subsection{Data Acquisition and Eligibility Criteria}
\label{sec:data_acquisition}

All data were collected at Shenzhen Eye Hospital between April and December 2023. The study was approved by the Ethics Committee of Shenzhen Eye Hospital (approval number: 2024KYPJ012) and was conducted in accordance with the Declaration of Helsinki, with written informed consent obtained from all participants.

Participants were classified into three refractive groups based on spherical equivalent (SE): emmetropia with SE from $+$2.0~D to $-$0.25~D, myopia with SE $\leq -$0.5~D, and high myopia with SE $\leq -$6.0~D~\cite{baird2020myopia}. Eyes were excluded if they had a history of vitreoretinal surgery or severe vitreoretinal diseases, including proliferative diabetic retinopathy, retinal vein or artery occlusion, retinitis pigmentosa, central serous chorioretinopathy, glaucoma, retinal inflammation, or optic nerve diseases.

Each participant underwent a comprehensive ophthalmic examination. Intraocular pressure (IOP) was measured using a non-contact tonometer (NT-530P, NIDEK). Refractive errors were evaluated with an autorefractor (PRK-7000, Potec) and refined by subjective refraction using a phoropter (VT-10, Topcon). Axial length, central corneal thickness (CCT), and keratometry were measured using an optical biometer (IOLMaster 700, Carl Zeiss Meditec).

\begin{figure*}[!t]
\centering
\includegraphics[width=\linewidth]{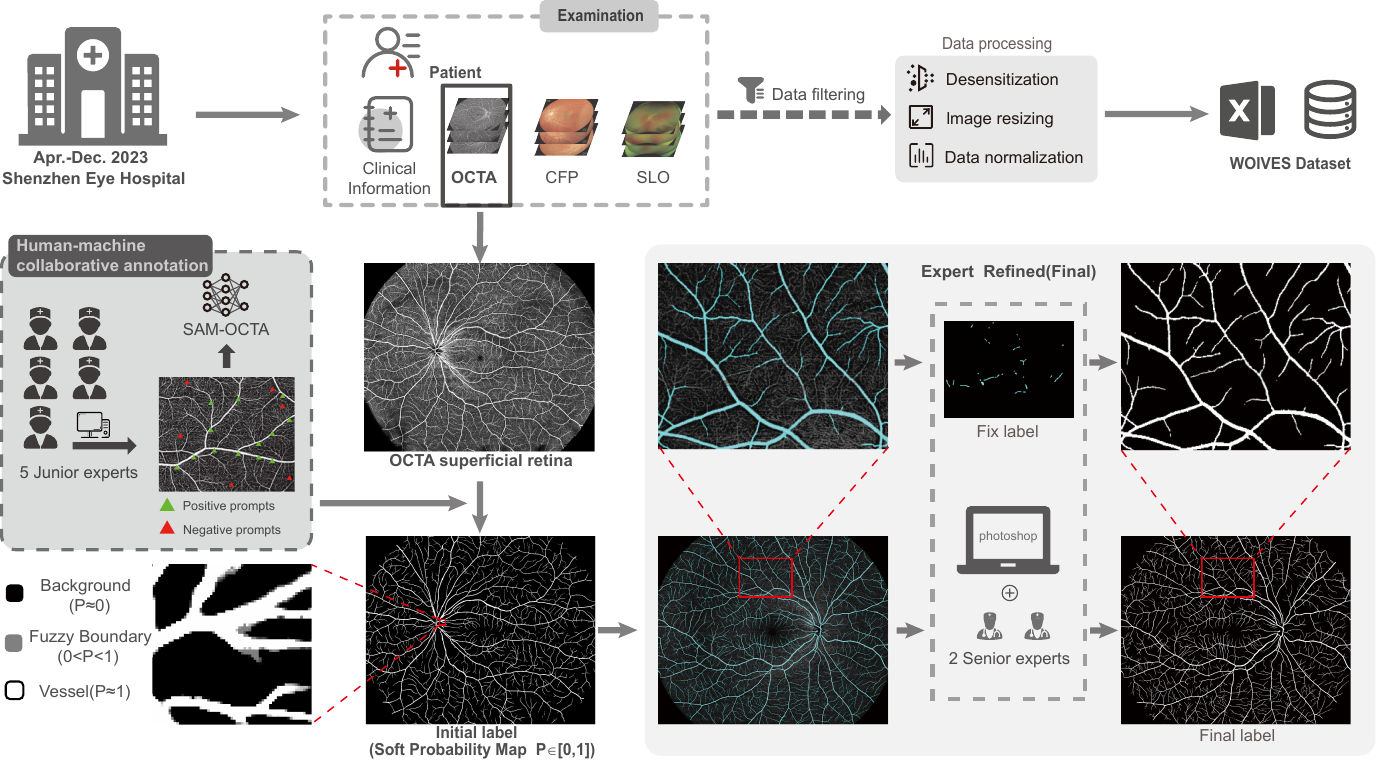}
\caption{Overview of the WOIVES dataset construction pipeline. Top: data acquisition at Shenzhen Eye Hospital (Apr.--Dec.\ 2023), including clinical information, OCTA, CFP, and SLO imaging, followed by data filtering, desensitization, image resizing, and normalization. Bottom left: human--machine collaborative annotation producing soft probability maps ($P \in [0, 1]$) through multi-annotator fusion. Bottom right: expert refinement producing the final vessel segmentation labels.}
\label{fig:pipeline}
\end{figure*}

Imaging was performed using a swept-source OCTA system (TowardPi Medical Technology, China), operating at an A-scan speed of 400{,}000 scans per second with an axial resolution of 3.8~$\mu$m and a lateral resolution of 10~$\mu$m. The maximum single-acquisition field of view (FOV) is 24$\times$20~mm, corresponding to a UWF fundus view of approximately 120\textdegree. For each eye, six types of angiographic and en face images were acquired: three OCTA images covering the superficial retina (from the internal limiting membrane to the outer plexiform layer), deep retina (from the outer plexiform layer to Bruch's membrane), and full retina (overlay of superficial and deep layers), along with three en face choroidal images of the choriocapillaris, choroidal medium and large vessels, and entire choroid. Color fundus photographs (CFP) were captured using a non-mydriatic fundus camera (SW8800, Suowei), and scanning laser ophthalmoscopy (SLO) images were obtained using an ultra-widefield imaging system (Optos Daytona). All OCTA images were stored in PNG format at 1536$\times$1280 pixels. The data acquisition and annotation workflow is illustrated in Fig.~\ref{fig:pipeline}.

\subsection{Data Processing}
\label{sec:data_processing}

Following acquisition, all images were manually reviewed, and those with significant quality issues, such as severe artifacts, defocus, motion blur, or exposure errors, were excluded. Retained images underwent desensitization to remove personally identifiable information, resizing to ensure uniform dimensions, and normalization to standardize pixel intensity distributions. Accompanying clinical metadata, including demographics, ocular biometry, refractive measurements, and diagnostic classifications, were organized in a structured table with all identifiable information removed.

\subsection{Annotation Pipeline}
\label{sec:annotation}

Vascular segmentation annotations were performed on the OCTA superficial retina images. Given the large spatial extent and dense vascular complexity of UWF OCTA images, we designed a two-stage human--machine collaborative annotation pipeline.

In the first stage, five junior ophthalmologists independently annotated each image with the assistance of SAM-OCTA~\cite{shi2024samocta}, an open-source interactive segmentation model adapted from the Segment Anything Model for OCTA vessel segmentation. Each annotator provided positive and negative prompts to guide the model, producing an initial binary segmentation mask. The five annotations were aggregated into a soft probability map $P \in [0, 1]$ by averaging across annotators, where values near~1 indicate high-confidence vessel regions, values near~0 indicate definitive background, and intermediate values encode boundary uncertainty.

In the second stage, two senior ophthalmology experts with over five years of clinical experience reviewed and refined the fused annotations by adding missed vessels, removing false positives, and correcting boundary delineations. The complete annotation pipeline is illustrated in Fig.~\ref{fig:pipeline}.

A distinguishing feature of this pipeline is that the final annotations are retained as soft probability maps rather than binarized into hard masks, following evidence that soft-label training improves both probability calibration and sensitivity to small structures relative to binary training~\cite{gros2021softseg}.
In UWF OCTA images, vessel boundaries, particularly those of the thinnest resolvable vessels in peripheral regions, are inherently ambiguous due to reduced signal-to-noise ratio and partial volume effects. % [REV] "fine capillaries" -> "thinnest resolvable vessels"
As illustrated in Fig.~\ref{fig:pipeline}, the fused probability map naturally decomposes each image into three categories: definitive vessel regions ($P \approx 1$), definitive background ($P \approx 0$), and uncertain boundary zones ($0 < P < 1$) where annotators disagreed. Conventional binary annotation forces a hard threshold that discards this disagreement, whereas our soft labels explicitly preserve it. This design directly affects model training: the composite loss (Section~\ref{sec:loss}) operates on continuous probability outputs, allowing the network to learn graded confidence rather than overcommitting at ambiguous boundaries. At inference time, the resulting soft-label-aware predictions provide finer-grained vessel boundary estimation that benefits the downstream vascular measurements evaluated in Section~\ref{sec:clinical_analysis}. % [REV] "probability-calibrated predictions" -> "soft-label-aware predictions"

\subsection{Dataset Statistics and Comparison}
\label{sec:dataset_stats}

\textbf{Cohort structure.}
WOIVES comprises two nested cohorts. The \emph{imaging cohort}, used for all segmentation experiments, cross-validation, and vascular measurement analyses, consists of 206 eyes from 152 participants with quality-controlled UWF OCTA images and expert-refined vessel annotations. Within this cohort, a \emph{clinically characterized subset} of 177 eyes from 130 participants additionally provides complete ophthalmic metadata, including axial length, SE, IOP, CCT, and refractive diagnosis. This nested organization distinguishes the full imaging cohort used for segmentation evaluation from the subset with complete clinical and biometric information.

\textbf{Comparison with existing datasets.}
Table~\ref{tab:dataset_comparison} compares WOIVES with existing public OCTA datasets. The 24$\times$20~mm$^2$ FOV (480~mm$^2$) is 13.3$\times$ larger than OCTA-500 (6$\times$6~mm$^2$, 36~mm$^2$) and 3.3$\times$ larger than DRAC (12$\times$12~mm$^2$, 144~mm$^2$). Compared with the public benchmarks considered here, WOIVES combines UWF SS-OCTA vessel annotations with full-depth choroidal imaging and ocular biometric information, including axial length measurements for the clinically characterized subset. The dataset is publicly available on Zenodo (DOI: 10.5281/zenodo.21904672).

\begin{table*}[!t]
\centering
\caption{Comparison of WOIVES with Representative Public OCTA Datasets}
\label{tab:dataset_comparison}
\resizebox{\linewidth}{!}{%
\begin{threeparttable}
\begin{tabular}{lccccccccc}
\toprule
Dataset & Year & Technology & FOV (mm\textsuperscript{2}) & Area & Size & Annotation & Disease Focus & Axial Length & Choroid \\
\midrule
OCTA-500~\cite{li2020octa500} & 2024 & SD-OCT (Optovue) & 6$\times$6 & 36 mm\textsuperscript{2} & 500 eyes & Vessel (binary) & DR, AMD, CNV & \xmark & \xmark \\
ROSE-1~\cite{ma2021rose}   & 2021 & SD-OCT (RTVue XR) & 3$\times$3 & 9 mm\textsuperscript{2} & 117 eyes & Vessel (binary) & Alzheimer's Disease & \xmark & \xmark \\
ROSE-2~\cite{ma2021rose}   & 2021 & SD-OCT (Heidelberg) & 3$\times$3 & 9 mm\textsuperscript{2} & 112 eyes & Vessel (binary) & Macular Disease & \xmark & \xmark \\
DRAC~\cite{shang2022drac}     & 2022 & SS-OCT (VG200D) & 12$\times$12 & 144 mm\textsuperscript{2} & 174 images\tnote{$\ddagger$} & DR lesion (binary) & Diabetic Retinopathy & \xmark & Partial \\
\midrule
WOIVES (Ours) & 2026 & SS-OCT (TowardPi) & 24$\times$20 & 480 mm\textsuperscript{2} & 206 eyes & \textbf{Vessel (soft prob.)} & High Myopia & \cmark & Full-depth \\
\bottomrule
\end{tabular}%
\end{threeparttable}%
}
\vspace{4pt}
% 独立的注释区域，宽度等于 \linewidth，自动换行
\noindent\scriptsize
\parbox{\linewidth}{%
\setlength{\itemsep}{0pt}%
\setlength{\parskip}{0pt}%
\setlength{\parsep}{0pt}%
\renewcommand{\baselinestretch}{0.90}\selectfont
\begin{itemize}
\item[$\ddagger$] DRAC comprises 1{,}103 UW-OCTA images in total; its lesion-segmentation task uses 109 training and 65 test images. Subject- and eye-level counts are not reported. DRAC annotations target DR lesions rather than the vascular tree, so it is not a vessel-segmentation benchmark.
\item[] WOIVES provides 13.3$\times$ larger FOV than OCTA-500 and 3.3$\times$ larger than DRAC. Abbreviations: DR: Diabetic Retinopathy; AMD: Age-related Macular Degeneration; CNV: Choroidal Neovascularization; SD-OCT: Spectral-Domain OCT; SS-OCT: Swept-Source OCT. Symbols: \cmark: Available; \xmark: Not available.
\end{itemize}
}
\end{table*}

\begin{comment}
\begin{table}[!t]
\centering
\caption{Clinical and biometric characteristics of the clinically characterized subset (177 eyes, 130 participants). Values are mean$\pm$SD; $p$ from Kruskal--Wallis tests across the three refractive groups.}
\label{tab:cohort}
\setlength{\tabcolsep}{2.5pt}
\renewcommand{\arraystretch}{1.15}
\resizebox{\linewidth}{!}{%
\begin{tabular}{lccccc}
\toprule
Characteristic & Emmetropia & Myopia & High Myopia & Overall & $p$ \\
\midrule
Eyes, $n$             & 30 & 76 & 71 & 177 & -- \\
Axial length (mm)     & 24.04$\pm$1.22 & 25.16$\pm$1.01 & 26.63$\pm$1.05 & 25.56$\pm$1.43 & $3{\times}10^{-18}$ \\
Spherical equiv.\ (D) & $+$0.77$\pm$1.20 & $-$2.87$\pm$1.57 & $-$7.09$\pm$1.35 & $-$3.95$\pm$3.21 & $6{\times}10^{-33}$ \\
CCT ($\mu$m)          & 517.1$\pm$56.7 & 526.6$\pm$42.0 & 540.8$\pm$33.0 & 530.7$\pm$42.4 & 0.031 \\
IOP (mmHg)            & 13.2$\pm$2.4 & 14.8$\pm$2.6 & 15.6$\pm$2.2 & 14.8$\pm$2.5 & $1{\times}10^{-4}$ \\
AL/K ratio            & 0.59$\pm$0.07 & 0.59$\pm$0.05 & 0.62$\pm$0.04 & 0.61$\pm$0.05 & $1{\times}10^{-6}$ \\
\midrule
\multicolumn{6}{l}{\footnotesize Participants: 130 (65 male, 65 female); age 33.8$\pm$5.7~yr (range 22--51).} \\
\multicolumn{6}{l}{\footnotesize Axial length range 22.30--29.11~mm. Family history of high myopia: 11 of 130.} \\
\bottomrule
\end{tabular}%
}
\end{table}
\end{comment}

% =====================================================================
\section{Method}
\label{sec:method}

This section presents PG-Mamba, a Mamba-based U-shaped network for UWF OCTA vessel segmentation. We review the preliminaries of state space models, describe the overall architecture, detail the two proposed modules, and present the training objective.
\subsection{Preliminaries: State Space Models}
\label{sec:ssm_prelim}

State space models (SSMs) map an input sequence $x(t) \in \mathbb{R}$ to an output $y(t) \in \mathbb{R}$ through a latent state $h(t) \in \mathbb{R}^{N}$. A continuous-time SSM is defined as
\begin{align}
    h'(t) &= \mathbf{A}h(t) + \mathbf{B}x(t), \\
    y(t) &= \mathbf{C}h(t),
\end{align}
where $\mathbf{A} \in \mathbb{R}^{N \times N}$ is the state transition matrix, $\mathbf{B} \in \mathbb{R}^{N \times 1}$ is the input projection matrix, and $\mathbf{C} \in \mathbb{R}^{1 \times N}$ is the output projection matrix.

For discrete computation, the continuous system can be discretized with step size $\Delta$ using the zero-order hold rule:
\begin{align}
    \overline{\mathbf{A}} &= \exp(\Delta\mathbf{A}), \\
    \overline{\mathbf{B}} &= (\Delta\mathbf{A})^{-1}
    \left(\exp(\Delta\mathbf{A})-\mathbf{I}\right)\Delta\mathbf{B},
\end{align}
which yields the recurrence
\begin{equation}
    h_k = \overline{\mathbf{A}}h_{k-1}
    + \overline{\mathbf{B}}x_k,
    \qquad
    y_k = \mathbf{C}h_k.
\end{equation}

Mamba~\cite{gu2023mamba} makes $\mathbf{B}$, $\mathbf{C}$, and $\Delta$ input-dependent, enabling selective state propagation with linear complexity in the sequence length. For two-dimensional visual features, the feature map must first be serialized into one-dimensional sequences for state-space processing~\cite{liu2024vmamba}. The scan order therefore determines how spatial locations are organized along the sequences over which state information is propagated, providing the basis for the Polar-Scan SS2D introduced below.

\subsection{Overall Architecture}
\label{sec:architecture}

\begin{figure*}[!t]
\centering
\includegraphics[width=\linewidth]{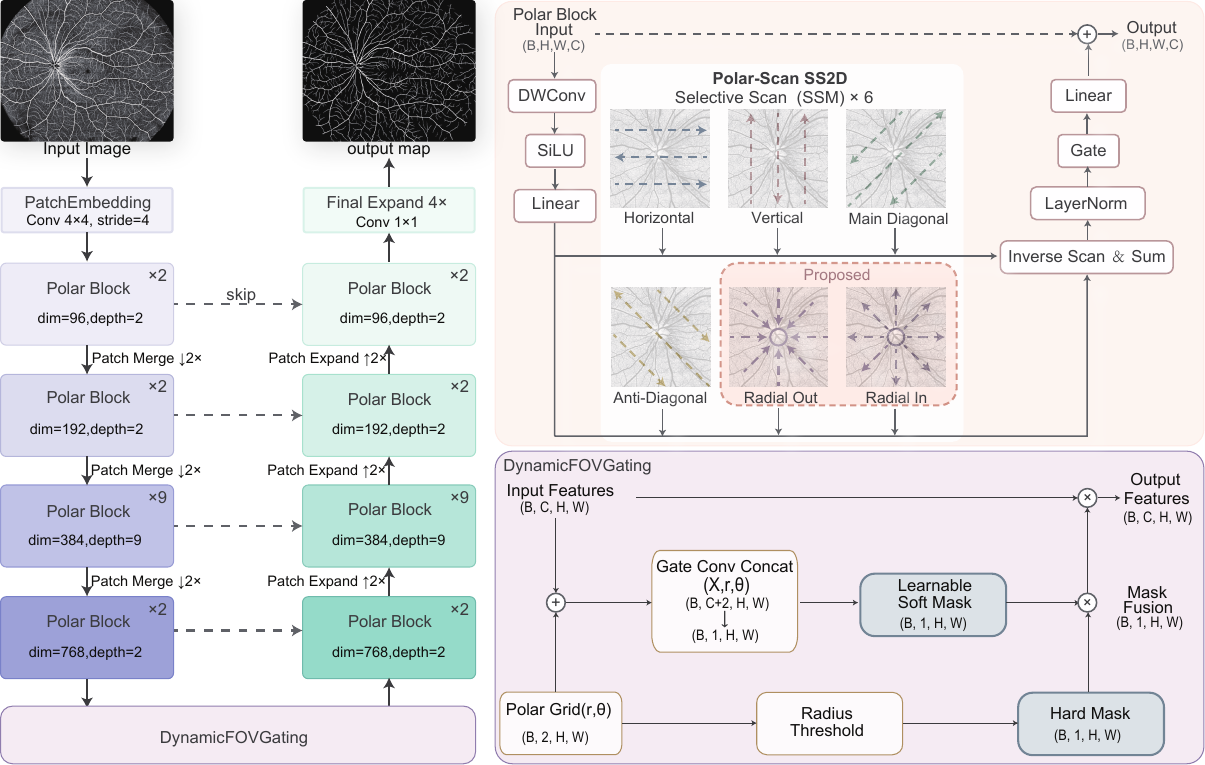}
\caption{Architecture of PG-Mamba. Left: the U-shaped encoder--decoder with four Polar Block stages, skip connections, and Dynamic FOV Gating at the bottleneck. Top right: the Polar Block with Polar-Scan SS2D using six scan orders, including the proposed Radial Out and Radial In scans. Bottom right: Dynamic FOV Gating combining a content-adaptive soft mask with a radius-thresholded hard mask.}
\label{fig:architecture}
\end{figure*}

PG-Mamba adopts a U-shaped encoder--decoder architecture for single-channel UWF OCTA vessel segmentation (Fig.~\ref{fig:architecture}). Given an input image $\mathbf{X} \in \mathbb{R}^{B \times 1 \times H \times W}$, the network predicts a vessel probability map $\hat{\mathbf{Y}} \in [0,1]^{B \times 1 \times H \times W}$. Polar-coordinate information required by the proposed modules is generated internally rather than provided as additional input channels.

A $4\times4$ convolution with stride 4 performs patch embedding and projects the input into a 96-dimensional feature space. The encoder contains four stages of Polar Blocks (Section~\ref{sec:polar_block}) with depths $[2,2,9,2]$ and channel dimensions $[96,192,384,768]$, with Patch Merge layers providing $2\times$ downsampling between adjacent stages. Dynamic FOV Gating (Section~\ref{sec:dynamic_gating}) is applied only to the bottleneck features. The decoder mirrors the encoder hierarchy using Polar Blocks, Patch Expand layers for $2\times$ upsampling, and additive skip connections from the corresponding encoder stages. A final $4\times$ patch expansion restores the original spatial resolution, followed by a $1\times1$ convolution and sigmoid activation to produce the full-resolution probability map. Both training and inference operate on $512\times512$ image tiles (Section~\ref{sec:exp_setup}), and the coordinates used by the proposed modules are computed internally on feature maps derived from the current tile.

\subsection{Polar Block and Polar-Scan SS2D}
\label{sec:polar_block}

The Polar Block builds on the visual state space (VSS) block~\cite{liu2024vmamba}, with Polar-Scan SS2D as its principal modification. Applying selective state-space models to images requires mapping two-dimensional features to ordered one-dimensional sequences. The scan order therefore determines the spatial ordering presented to the selective scan operator. Polar-Scan SS2D extends this mapping by introducing two polar-coordinate-based scan orders in addition to the conventional directional scans.

Given an input feature tensor $\mathbf{F} \in \mathbb{R}^{B \times H \times W \times C}$, the Polar Block retains the gated residual structure of the VSS block while replacing its selective scanning operation with Polar-Scan SS2D:
\begin{equation}
\begin{aligned}
    [\mathbf{X},\mathbf{Z}]
    &=
    \operatorname{Split}
    \left(
    \operatorname{Linear}_{\mathrm{in}}
    \left(
    \operatorname{LN}(\mathbf{F})
    \right)
    \right), \\
    \mathbf{Y}
    &=
    \operatorname{PolarScanSS2D}
    \left(
    \operatorname{SiLU}
    \left(
    \operatorname{DWConv}_{3\times3}(\mathbf{X})
    \right)
    \right).
\end{aligned}
\label{eq:polar_block}
\end{equation}
The block output is then obtained through gated projection and residual connection:
\begin{equation}
    \mathbf{F}_{\mathrm{out}}
    =
    \mathbf{F}
    +
    \operatorname{DropPath}
    \left[
    \operatorname{Linear}_{\mathrm{out}}
    \left(
    \operatorname{LN}(\mathbf{Y})
    \odot
    \operatorname{SiLU}(\mathbf{Z})
    \right)
    \right].
    \label{eq:polar_block_output}
\end{equation}

Polar-Scan SS2D augments the conventional directional scans with two additional polar scan orders. For a feature map of size $H\times W$, the local polar coordinates of position $(i,j)$ are defined as
\begin{equation}
    r_{i,j}
    =
    \sqrt{(i-c_h)^2+(j-c_w)^2},
    \qquad
    \theta_{i,j}
    =
    \operatorname{atan2}(i-c_h,\,j-c_w),
    \label{eq:polar_coordinates}
\end{equation}
where $c_h=\lfloor H/2 \rfloor$ and $c_w=\lfloor W/2 \rfloor$ denote the feature-map center.

Let $L=HW$. The Radial Out and Radial In scan orders are constructed using the polar scores
\begin{align}
    q_{\mathrm{RO}}(i,j)
    &=
    (L+1)r_{i,j}
    +
    \lambda_{\theta}(\theta_{i,j}+\pi), \\
    q_{\mathrm{RI}}(i,j)
    &=
    -(L+1)r_{i,j}
    +
    \lambda_{\theta}(\theta_{i,j}+\pi),
    \label{eq:polar_scores}
\end{align}
where $\lambda_{\theta}=10$ is fixed throughout the experiments. Sorting spatial positions according to these scores produces the Radial Out and Radial In permutations, respectively. The opposite signs of the radial term induce opposite dominant radial orderings, while the angular term jointly determines the traversal within each polar sequence.

Together with the horizontal, vertical, main-diagonal, and anti-diagonal scans, the complete scan set is
\begin{equation}
    \mathcal{S}
    =
    \{
    s_{\mathrm{H}},
    s_{\mathrm{V}},
    s_{\mathrm{MD}},
    s_{\mathrm{AD}},
    s_{\mathrm{RO}},
    s_{\mathrm{RI}}
    \}.
    \label{eq:scan_set}
\end{equation}
Each scan route is processed by its corresponding selective state-space parameters and then restored to the original spatial arrangement through the inverse permutation. The six restored feature maps are finally aggregated by direct summation:
\begin{equation}
    \operatorname{PolarScanSS2D}(\mathbf{X})
    =
    \sum_{s\in\mathcal{S}}
    \operatorname{InverseScan}_{s}
    \left(
    \operatorname{SSM}_{s}
    \left(
    \operatorname{Scan}_{s}(\mathbf{X})
    \right)
    \right).
    \label{eq:polar_scan}
\end{equation}

The two proposed polar scans complement the four directional scans by providing additional spatial orderings determined jointly by radial and angular position.

\subsection{Dynamic FOV Gating}
\label{sec:dynamic_gating}

Dynamic FOV Gating is introduced as an auxiliary spatial modulation module at the encoder--decoder bottleneck. Given bottleneck features
$\mathbf{F} \in \mathbb{R}^{B \times C \times H \times W}$,
we construct a normalized polar grid
$\mathbf{G}=[\widetilde{\mathbf{R}};\widetilde{\mathbf{\Theta}}]
\in \mathbb{R}^{B \times 2 \times H \times W}$.
For each spatial position $(i,j)$, the normalized Cartesian coordinates are defined as
\begin{equation}
    x_{i,j}
    =
    2\frac{j}{W-1}-1,
    \qquad
    y_{i,j}
    =
    2\frac{i}{H-1}-1,
\end{equation}
from which the corresponding polar coordinates are obtained as
\begin{equation}
    \tilde{r}_{i,j}
    =
    \sqrt{x_{i,j}^{2}+y_{i,j}^{2}},
    \qquad
    \tilde{\theta}_{i,j}
    =
    \frac{\operatorname{atan2}(y_{i,j},x_{i,j})}{\pi}.
    \label{eq:fov_coordinates}
\end{equation}

The polar grid is concatenated with the bottleneck features to generate a content-adaptive soft mask:
\begin{equation}
    \mathbf{M}_{\mathrm{soft}}
    =
    \sigma
    \left(
    g_{\phi}
    \left(
    [\mathbf{F};\mathbf{G}]
    \right)
    \right),
    \label{eq:soft_mask}
\end{equation}
where $\sigma$ denotes the sigmoid function and $g_{\phi}$ consists of two $3\times3$ convolutions with GroupNorm and ReLU between them. In parallel, a fixed radial support mask is defined as
\begin{equation}
    \mathbf{M}_{\mathrm{hard}}(i,j)
    =
    \mathbb{I}
    \left(
    \tilde{r}_{i,j}\leq\tau
    \right),
    \qquad
    \tau=1.15.
    \label{eq:hard_mask}
\end{equation}
The gated bottleneck features are then obtained by
\begin{equation}
    \mathbf{F}_{\mathrm{out}}
    =
    \mathbf{F}
    \odot
    \mathbf{M}_{\mathrm{soft}}
    \odot
    \mathbf{M}_{\mathrm{hard}}.
    \label{eq:fov_gating}
\end{equation}

Both masks are spatial and shared across channels. The soft mask provides feature-dependent weighting, while the hard mask imposes a fixed radial constraint on the bottleneck features.

\subsection{Loss Function}
\label{sec:loss}

Because the WOIVES annotations are retained as soft probability maps, we use a composite loss combining soft Dice loss and mean squared error:
\begin{equation}
    \mathcal{L}
    =
    \frac{1}{2}
    \left(
    1 -
    \frac{2\sum_{i} p_i g_i + \epsilon}
    {\sum_{i} p_i + \sum_{i} g_i + \epsilon}
    \right)
    +
    \frac{1}{2N}
    \sum_{i=1}^{N}
    (p_i-g_i)^2,
    \label{eq:total_loss}
\end{equation}
where $p_i$ denotes the predicted vessel probability, $g_i \in [0,1]$ is the corresponding soft annotation value from the probability map, $N$ is the number of pixels, and $\epsilon$ is a smoothing constant. The soft Dice term promotes spatial overlap, while the mean squared error provides pixel-wise supervision for the continuous annotations. All compared methods are trained using the same loss function.

% =====================================================================
\section{Experiments}
\label{sec:experiments}

\subsection{Experimental Setup}
\label{sec:exp_setup}

\subsubsection{Dataset and Cross-Validation}
All experiments were conducted on the WOIVES imaging cohort (206 eyes from 152 participants) using superficial-retina OCTA images. We adopted subject-level five-fold cross-validation with a \texttt{GroupKFold} partition, ensuring that both eyes from the same participant were assigned to the same fold and that no participant appeared in both the training and test sets. Within each fold, a subject-disjoint validation split was held out from the training portion for model selection. Unless otherwise stated, the reported values are the mean$\pm$standard deviation across the five test folds. For the eye-level paired analyses reported below, per-eye scores were pooled across the five test folds ($n=206$). Because some participants contributed both eyes, a participant-level sensitivity analysis, in which paired differences were averaged within participants before statistical summarization, is provided in the supplementary material.

\subsubsection{Evaluation Metrics}
We evaluated segmentation performance at overlap, probability, and topology levels. \emph{Dice} and \emph{Intersection over Union (IoU)} measure pixel-wise overlap between thresholded predictions and binarized reference annotations. \emph{Soft Dice} operates directly on the continuous prediction maps and soft labels, measuring agreement with the continuous reference without thresholding. \emph{Mean absolute error (MAE)} and the \emph{Brier score}~\cite{brier1950verification} quantify pixel-wise error on the continuous maps. MAE accumulates absolute deviations linearly, whereas the Brier score gives greater weight to larger individual errors. \emph{Centerline Dice (clDice)}~\cite{shit2021cldice} evaluates agreement between skeletonized vessel structures and is therefore sensitive to vascular continuity. All models produce probability maps in $[0,1]$. A fixed threshold of 0.5 was used for binary metrics, without model-specific threshold tuning.

\subsubsection{Implementation Details}
All models were implemented in PyTorch and trained on a single NVIDIA RTX 3090 GPU. Training used 512$\times$512 image tiles. Full-resolution inference was performed with a 512$\times$512 sliding window and 50\% overlap, with overlapping probability predictions averaged. Data augmentation included random flipping, rotation, and cropping. We used the AdamW optimizer with cosine-annealed learning rates and a weight decay of $10^{-2}$. The initial learning rate was $3\times10^{-4}$ for PG-Mamba and the Transformer- and Mamba-based models, and $1\times10^{-3}$ for the CNN-based baselines. Early stopping and model selection were performed on the subject-disjoint validation split. All compared methods were trained using the same composite loss described in Section~\ref{sec:loss}, and a fixed random seed was used throughout the experiments.

\subsubsection{Compared Methods}
We compared PG-Mamba with seven segmentation methods spanning three architecture families: CNN-based methods, including U-Net~\cite{ronneberger2015unet}, UNet++~\cite{zhou2018unetpp}, and R2U-Net~\cite{alom2018r2unet}; Transformer-based methods, including Swin-UNet~\cite{cao2022swinunet} and H2Former~\cite{he2023h2former}; and Mamba-based methods, including VM-UNet~\cite{ruan2024vmunet} and AC-MambaSeg~\cite{yang2024acmamba}. AC-MambaSeg was originally proposed for skin-lesion segmentation. It and all other baselines were retrained on WOIVES using the same single-channel input and composite loss.

\subsection{Comparative Segmentation Performance}
\label{sec:comparison}

Table~\ref{tab:model_comparison} summarizes the quantitative comparison on WOIVES. PG-Mamba achieves the highest performance on five of the six reported metrics, including Dice (89.81\%), IoU (82.20\%), soft Dice (85.99\%), clDice (91.97\%), and MAE (0.0257). The only exception is the Brier score, for which H2Former obtains the lowest value (0.0146), while PG-Mamba and AC-MambaSeg share the second-lowest value (0.0156).

The improvement over the strongest overall baseline, VM-UNet, is modest in magnitude: PG-Mamba improves the five-fold mean Dice by 0.44 percentage points and clDice by 0.38 percentage points. Eye-level paired Wilcoxon signed-rank tests across the pooled test predictions ($n=206$ eyes) indicate differences between PG-Mamba and VM-UNet for Dice, IoU, soft Dice, MAE, Brier, and clDice (all $p<10^{-12}$). PG-Mamba also differs significantly from the remaining baselines on the overlap and topology metrics and on soft Dice and MAE (all $p<10^{-3}$). The exceptions are confined to the Brier score, for which H2Former performs better ($p=5\times10^{-4}$), while the difference between AC-MambaSeg and PG-Mamba was not statistically significant. Because these eye-level tests do not fully account for bilateral correlation, the supplementary participant-level analysis is used as a sensitivity check rather than interpreting the small $p$-values as evidence of a large practical effect.

The probability-level metrics provide complementary information. AC-MambaSeg obtains a Brier score comparable to that of PG-Mamba but records substantially lower soft Dice and higher MAE, whereas H2Former achieves the lowest Brier score without matching PG-Mamba on overlap or topology. The Brier score alone therefore does not fully characterize agreement with the continuous vessel annotations in this sparse-foreground setting. Among the evaluated methods, PG-Mamba provides the most balanced performance across binary overlap, soft-label agreement, pixel-wise error, and vascular connectivity.

\begin{table*}[!t]
\centering
\caption{Segmentation performance on WOIVES under subject-level five-fold cross-validation (mean$\pm$standard deviation). Best values are shown in bold and second-best values are underlined.}
\label{tab:model_comparison}
\setlength{\tabcolsep}{5pt}
\renewcommand{\arraystretch}{1.2}
\begin{tabular}{llcccccc}
\toprule
Backbone & Model & Dice (\%) & IoU (\%) & Soft Dice (\%) & clDice (\%) & MAE & Brier \\
\midrule
\multirow{3}{*}{CNN}
 & U-Net    & 87.78\tiny{$\pm$1.49} & 78.80\tiny{$\pm$2.18} & 83.31\tiny{$\pm$1.44} & 90.76\tiny{$\pm$1.19} & 0.0288\tiny{$\pm$0.0027} & 0.0166\tiny{$\pm$0.0022} \\
 & UNet++  & 88.92\tiny{$\pm$0.95} & 80.68\tiny{$\pm$1.36} & 85.07\tiny{$\pm$0.86} & 91.47\tiny{$\pm$0.94} & \underline{0.0265}\tiny{$\pm$0.0014} & 0.0158\tiny{$\pm$0.0013} \\
 & R2U-Net & 85.97\tiny{$\pm$1.73} & 75.98\tiny{$\pm$2.80} & 82.40\tiny{$\pm$1.58} & 88.18\tiny{$\pm$2.00} & 0.0337\tiny{$\pm$0.0039} & 0.0219\tiny{$\pm$0.0030} \\
\midrule
\multirow{2}{*}{Transformer}
 & Swin-UNet & 89.04\tiny{$\pm$0.64} & 80.90\tiny{$\pm$0.79} & 85.34\tiny{$\pm$0.59} & 90.96\tiny{$\pm$0.84} & 0.0269\tiny{$\pm$0.0010} & 0.0168\tiny{$\pm$0.0010} \\
 & H2Former  & 88.02\tiny{$\pm$1.02} & 79.17\tiny{$\pm$1.45} & 82.44\tiny{$\pm$2.02} & 91.42\tiny{$\pm$1.09} & 0.0283\tiny{$\pm$0.0030} & \textbf{0.0146}\tiny{$\pm$0.0017} \\
\midrule
\multirow{3}{*}{Mamba}
 & VM-UNet     & \underline{89.37}\tiny{$\pm$0.53} & \underline{81.45}\tiny{$\pm$0.59} & \underline{85.61}\tiny{$\pm$0.52} & \underline{91.59}\tiny{$\pm$0.72} & \underline{0.0265}\tiny{$\pm$0.0010} & 0.0163\tiny{$\pm$0.0010} \\
 & AC-MambaSeg & 88.49\tiny{$\pm$1.32} & 79.97\tiny{$\pm$1.91} & 75.76\tiny{$\pm$4.29} & 90.95\tiny{$\pm$1.20} & 0.0464\tiny{$\pm$0.0103} & \underline{0.0156}\tiny{$\pm$0.0021} \\
 & PG-Mamba    & \textbf{89.81}\tiny{$\pm$0.92} & \textbf{82.20}\tiny{$\pm$1.23} & \textbf{85.99}\tiny{$\pm$0.82} & \textbf{91.97}\tiny{$\pm$1.01} & \textbf{0.0257}\tiny{$\pm$0.0015} & \underline{0.0156}\tiny{$\pm$0.0015} \\
\bottomrule
\end{tabular}
\end{table*}

\subsection{Ablation Study}
\label{sec:ablation}

We evaluated the contributions of Polar Scan and Dynamic FOV Gating using the same subject-disjoint five-fold protocol. Starting from the full PG-Mamba, we first removed Dynamic FOV Gating while retaining all six scan branches ($-$Gate). We then removed the radial-outward and radial-inward branches, leaving the four horizontal, vertical, main-diagonal, and anti-diagonal scans ($-$Polar). This sequential design allows the incremental effects of the two proposed components to be examined within the same PG-Mamba framework. Table~\ref{tab:ablation} summarizes the results.

\emph{Effect of Polar Scan.}
Adding the radial-outward and radial-inward branches ($-$Polar$\rightarrow$$-$Gate) improves the five-fold mean Dice from 89.43\% to 89.67\% and clDice from 91.54\% to 91.73\%, while reducing the Brier score from 0.0162 to 0.0158. The improvement in Dice is also supported by the eye-level paired analysis ($p=1.3\times10^{-7}$). These results show that incorporating the two polar scan branches provides a modest but consistent improvement over the four-direction configuration.

\emph{Effect of Dynamic FOV Gating.}
Adding Dynamic FOV Gating to the six-direction configuration ($-$Gate$\rightarrow$full) further improves Dice from 89.67\% to 89.81\% and clDice from 91.73\% to 91.97\%, while further decreasing the Brier score from 0.0158 to 0.0156. The improvements are significant in the eye-level paired analysis for both Dice ($p=1.2\times10^{-5}$) and clDice ($p=1.1\times10^{-7}$). The gating module therefore provides an additional improvement in both segmentation overlap and vascular structural agreement.

Overall, the controlled ablations exhibit a progressive improvement from the four-direction configuration to the six-direction Polar Scan configuration and finally to the full PG-Mamba. The results support complementary contributions from Polar Scan and Dynamic FOV Gating, with the complete model achieving the best Dice, clDice, and Brier score among the controlled PG-Mamba configurations.
\begin{table}[!t]
\centering
\caption{Ablation of Polar Scan and Dynamic FOV Gating under subject-level five-fold cross-validation.}
\label{tab:ablation}
\setlength{\tabcolsep}{4pt}
\renewcommand{\arraystretch}{1.2}
\resizebox{\linewidth}{!}{%
\begin{tabular}{lccc}
\toprule
Configuration & Dice (\%) & clDice (\%) & Brier \\
\midrule
VM-UNet (reference)
& 89.37\tiny{$\pm$0.53}
& 91.59\tiny{$\pm$0.72}
& 0.0163\tiny{$\pm$0.0010} \\
\midrule
$-$Polar ($-$Gate, 4 dir.)
& 89.43\tiny{$\pm$1.01}
& 91.54\tiny{$\pm$1.21}
& 0.0162\tiny{$\pm$0.0018} \\

$-$Gate (6 dir., no gate)
& 89.67\tiny{$\pm$0.93}
& 91.73\tiny{$\pm$1.13}
& 0.0158\tiny{$\pm$0.0016} \\

PG-Mamba (full)
& \textbf{89.81}\tiny{$\pm$0.92}
& \textbf{91.97}\tiny{$\pm$1.01}
& \textbf{0.0156}\tiny{$\pm$0.0015} \\
\bottomrule
\end{tabular}%
}

\begin{flushleft}
\footnotesize
The controlled variants were implemented within the PG-Mamba framework by disabling the corresponding components. The $-$Polar configuration is close to, but not identical to, the official VM-UNet implementation. Eye-level paired tests compare adjacent configurations within the controlled PG-Mamba block ($n=206$): adding the polar scan branches, Dice $p=1.3\times10^{-7}$; adding Dynamic FOV Gating, Dice $p=1.2\times10^{-5}$ and clDice $p=1.1\times10^{-7}$.
\end{flushleft}
\end{table}

\subsection{Vascular Measurement Analysis}
\label{sec:clinical_analysis}

We further evaluated whether differences in segmentation performance were reflected in downstream quantitative vascular measurements. Four complementary measurements were derived from each binarized prediction and its corresponding annotation-derived reference~\cite{yao2020quantitative}: vessel density (VD), quantified as the fraction of vessel pixels within the valid imaging region; fractal dimension (FD), estimated using box counting within the valid FOV; normalized vessel length density (VLD), quantified as the number of skeletonized vessel pixels per $10^3$ FOV pixels; and the normalized connected-component count (NCC), defined as the number of 8-connected components per $10^5$ FOV pixels and used as a connectivity-sensitive proxy. Predictions and soft annotations were binarized at a fixed threshold of 0.5 before measurement extraction. For each eye, the absolute error between the prediction-derived and annotation-derived measurements was calculated. PG-Mamba was compared with U-Net, R2U-Net, and VM-UNet using eye-level paired Wilcoxon signed-rank tests, with Holm correction applied across the 12 vascular-measurement comparisons.

\begin{figure}[!t]
\centering
\includegraphics[width=\columnwidth]{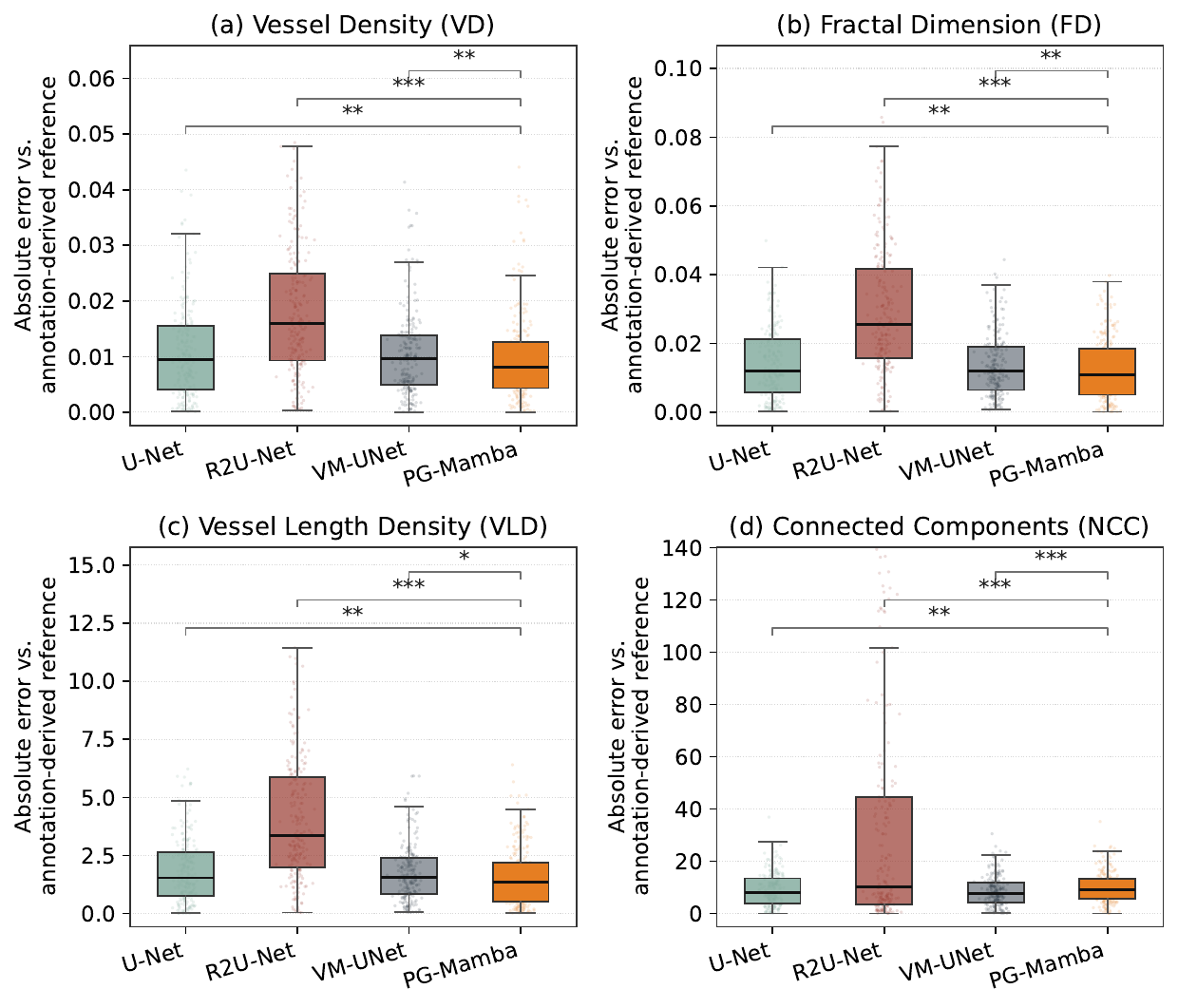}
\caption{Absolute errors of four vascular measurements derived from out-of-fold predictions for all 206 eyes. Errors are calculated relative to the corresponding annotation-derived measurements, with lower values indicating better agreement. VLD is expressed as skeletonized vessel pixels per $10^3$ FOV pixels, and NCC as the number of 8-connected components per $10^5$ FOV pixels. Brackets indicate eye-level paired Wilcoxon signed-rank tests comparing the absolute errors of PG-Mamba with each baseline after Holm correction ($^{*}p_{\mathrm{Holm}}<0.05$, $^{**}p_{\mathrm{Holm}}<0.01$, $^{***}p_{\mathrm{Holm}}<0.001$; n.s., not significant).}
\label{fig:biomarker}
\end{figure}

As shown in Fig.~\ref{fig:biomarker}, PG-Mamba achieved the lowest median absolute errors for VD, FD, and VLD, with values of 0.0081, 0.0109, and 1.3414, respectively. The corresponding median errors were 0.0095, 0.0121, and 1.5393 for U-Net; 0.0160, 0.0256, and 3.3652 for R2U-Net; and 0.0096, 0.0121, and 1.5606 for VM-UNet. For all three measurements, the paired absolute-error differences between PG-Mamba and each comparison model remained statistically significant after Holm correction ($p_{\mathrm{Holm}}<0.05$).

NCC exhibited a different pattern. PG-Mamba obtained a median absolute error of 9.2508, compared with 7.9610 for U-Net, 10.0523 for R2U-Net, and 7.7960 for VM-UNet. Paired analysis showed that PG-Mamba produced significantly lower NCC errors than R2U-Net ($p_{\mathrm{Holm}}<0.001$), but significantly higher errors than U-Net and VM-UNet ($p_{\mathrm{Holm}}<0.01$ and $p_{\mathrm{Holm}}<0.001$, respectively). Thus, the advantage of PG-Mamba in VD, FD, and VLD did not extend uniformly to this connectivity-sensitive measurement.

These findings also illustrate that different vascular measurements respond to different properties of a segmentation. VD primarily reflects the overall segmented vessel area, whereas FD summarizes geometric complexity and VLD depends on the extent of the skeletonized vascular network. NCC is particularly sensitive to small disconnected regions and local breaks, because changes in connectivity can directly alter the number of detected components. The results therefore indicate that improved segmentation performance can be accompanied by lower errors in several downstream vascular measurements, but does not necessarily imply superior agreement for every connectivity-sensitive quantity. Importantly, the present analysis evaluates agreement with measurements derived from the reference annotations and should not be interpreted as independent validation of clinical measurement accuracy.

% =====================================================================
\section{Discussion}
\label{sec:discussion}

\textbf{A dataset for myopia-oriented UWF vascular analysis.}
WOIVES addresses a gap in public resources for large-field OCTA vessel segmentation. Existing public OCTA resources used for segmentation are based on substantially smaller fields of view and were developed for different disease or analysis settings. Compared with conventional public benchmarks, the 24$\times$20~mm$^2$ (480~mm$^2$) acquisition in WOIVES provides a substantially larger imaging area, creating a large-field segmentation setting with dense vascular structures and spatially heterogeneous image quality. The clinically characterized subset further provides ocular biometric measurements across emmetropia, myopia, and high myopia. To our knowledge, this combination of UWF SS-OCTA vessel annotations, soft probability maps, and ocular biometry is not available in existing public OCTA vessel-segmentation benchmarks. Although the present study focuses on vessel segmentation, these characteristics may also support future investigation of myopia-related vascular changes and multimodal ophthalmic analysis.

\textbf{Why soft labels and multi-level evaluation matter.}
The continuous annotations in WOIVES enable evaluation beyond threshold-based Dice and IoU. Soft Dice, MAE, and Brier capture complementary aspects of agreement with the soft vessel annotations, and clDice provides an additional measure of vascular structural consistency. The rankings across these metrics are not identical. H2Former obtains the lowest Brier score without achieving the best overlap or topology performance, whereas AC-MambaSeg obtains a Brier score comparable to that of PG-Mamba but substantially lower soft Dice and higher MAE. These results show that no single probability-level metric fully characterizes segmentation quality in this sparse-foreground setting. A supplementary local-thickness analysis further shows that higher recall in the small local-thickness region does not necessarily correspond to better overall segmentation performance. Together, these findings support multi-level evaluation using complementary overlap, probability-sensitive, and topology-aware metrics.

\textbf{From segmentation to vascular measurement.}
The vascular measurement analysis was designed to assess whether segmentation differences were reflected in downstream quantitative agreement rather than to establish clinical validity. PG-Mamba achieved the lowest median absolute errors for VD, FD, and VLD, and the differences from all three comparison models remained significant after Holm correction. NCC showed a different pattern: PG-Mamba had lower error than R2U-Net but higher error than U-Net and VM-UNet. This divergence is consistent with the different structural properties emphasized by these measurements. VD primarily reflects segmented vessel area, FD summarizes geometric complexity, VLD depends on the extent of the skeletonized vascular network, and NCC is particularly sensitive to changes in connected-component structure. Accordingly, improved segmentation performance can be accompanied by improved agreement in several downstream vascular measurements without guaranteeing uniform improvement across every connectivity-sensitive quantity. This result also defines the scope of the present evidence more clearly: PG-Mamba improves several annotation-derived vascular measurements, but does not provide uniform superiority for all derived quantities.

\textbf{Limitations.}
Several limitations should be acknowledged. First, WOIVES was collected at a single center using a single SS-OCTA device, and the cohort contained more myopic and highly myopic eyes than emmetropic eyes. Multi-center and multi-device studies with more balanced populations are therefore needed to evaluate external validity and cross-device generalization. Second, the current annotation pipeline focuses on superficial retinal vessels. The soft probability maps represent consensus information derived from multiple annotations and should not be interpreted as calibrated biological probabilities. Extending the annotation framework to deeper retinal and choroidal vascular layers would enable more comprehensive layer-specific analysis. Third, the current PG-Mamba design uses fixed polar scan construction and applies Dynamic FOV Gating only at the bottleneck. Future work could investigate more adaptive scan-order strategies and multi-scale spatial modulation to further improve segmentation across heterogeneous vascular patterns. Finally, the vascular measurements were evaluated against references derived from expert annotations and consequently assess agreement with the annotation-derived reference rather than independent clinical validity. Future studies could incorporate longitudinal outcomes, device-derived quantitative measurements, and additional vascular measurements such as foveal avascular zone and regional perfusion metrics to assess broader clinical utility.

% ====================================================================
\section{Conclusion}
\label{sec:conclusion}

This work introduced WOIVES, a public UWF SS-OCTA vessel-segmentation dataset, together with PG-Mamba for large-field retinal vessel segmentation. PG-Mamba combines two complementary polar-coordinate scan orders with Dynamic FOV Gating for spatial feature modeling. Under subject-level five-fold cross-validation, PG-Mamba elevates segmentation performance across five of the six primary metrics, with modest but consistent improvements over strong CNN, Transformer, and Mamba baselines. Downstream vascular analysis further showed the lowest annotation-derived errors for VD, FD, and VLD. Together, WOIVES and PG-Mamba provide a benchmark and methodological basis for further research in UWF OCTA vessel segmentation and quantitative retinal vascular analysis.

% =====================================================================
%  REFERENCES  (37 entries)
% =====================================================================


\begin{thebibliography}{00}

% 1
\bibitem{holden2016global}
B.~A. Holden \emph{et al.},
``Global prevalence of myopia and high myopia and temporal trends from 2000 through 2050,''
\emph{Ophthalmology},
vol.~123, no.~5, pp.~1036--1042, May 2016,
doi: 10.1016/j.ophtha.2016.01.006.

% 2
\bibitem{baird2020myopia}
P.~N. Baird \emph{et al.},
``Myopia,''
\emph{Nat. Rev. Dis. Primers},
vol.~6, no.~1, Art.~no.~99, 2020,
doi: 10.1038/s41572-020-00231-4.

% 3
\bibitem{neelam2012choroidal}
K.~Neelam, C.~M.~G. Cheung, K.~Ohno-Matsui, T.~Y.~Y. Lai, and T.~Y. Wong,
``Choroidal neovascularization in pathological myopia,''
\emph{Prog. Retin. Eye Res.},
vol.~31, no.~5, pp.~495--525, Sep. 2012,
doi: 10.1016/j.preteyeres.2012.04.001.

% 4
\bibitem{he2019association}
J.~He \emph{et al.},
``Association between retinal microvasculature and optic disc alterations in high myopia,''
\emph{Eye},
vol.~33, no.~9, pp.~1494--1503, Sep. 2019,
doi: 10.1038/s41433-019-0438-7.

% 5
\bibitem{flores2013relationship}
I.~Flores-Moreno, F.~Lugo, J.~S. Duker, and J.~M. Ruiz-Moreno,
``The relationship between axial length and choroidal thickness in eyes with high myopia,''
\emph{Amer. J. Ophthalmol.},
vol.~155, no.~2, pp.~314--319.e1, Feb. 2013,
doi: 10.1016/j.ajo.2012.07.015.

% 6
\bibitem{liu2025uwfmyopia}
Y.~Liu \emph{et al.},
``Diagnosing pathologic myopia by identifying morphologic patterns using ultra widefield images with deep learning,''
\emph{npj Digit. Med.},
vol.~8, Art.~no.~435, 2025,
doi: 10.1038/s41746-025-01849-y.

% 7
\bibitem{invernizzi2020imaging}
A.~Invernizzi \emph{et al.},
``Imaging the choroid: From indocyanine green angiography to optical coherence tomography angiography,''
\emph{Asia-Pacific J. Ophthalmol.},
vol.~9, no.~4, pp.~335--348, Jul.--Aug. 2020,
doi: 10.1097/APO.0000000000000307.

% 8
\bibitem{sampson2022towards}
D.~M. Sampson, A.~M. Dubis, F.~K. Chen, R.~J. Zawadzki, and D.~D. Sampson,
``Towards standardizing retinal optical coherence tomography angiography: A review,''
\emph{Light: Sci. Appl.},
vol.~11, Art.~no.~63, 2022,
doi: 10.1038/s41377-022-00740-9.

% 9
\bibitem{zheng2023advances}
F.~Zheng \emph{et al.},
``Advances in swept-source optical coherence tomography and optical coherence tomography angiography,''
\emph{Adv. Ophthalmol. Pract. Res.},
vol.~3, no.~2, pp.~67--79, 2023,
doi: 10.1016/j.aopr.2022.10.005.

% 10
\bibitem{li2017retinal}
M.~Li \emph{et al.},
``Retinal microvascular network and microcirculation assessments in high myopia,''
\emph{Amer. J. Ophthalmol.},
vol.~174, pp.~56--67, Feb. 2017,
doi: 10.1016/j.ajo.2016.10.018.

% 11
\bibitem{yao2020quantitative}
X.~Yao, M.~N. Alam, D.~Le, and D.~Toslak,
``Quantitative optical coherence tomography angiography: A review,''
\emph{Exp. Biol. Med.},
vol.~245, no.~4, pp.~301--312, Feb. 2020,
doi: 10.1177/1535370219899893.

% 12
\bibitem{li2020octa500}
M.~Li \emph{et al.},
``OCTA-500: A retinal dataset for optical coherence tomography angiography study,''
\emph{Med. Image Anal.},
vol.~93, Art.~no.~103092, Apr. 2024,
doi: 10.1016/j.media.2024.103092.

% 13
\bibitem{ma2021rose}
Y.~Ma \emph{et al.},
``ROSE: A retinal OCT-angiography vessel segmentation dataset and new model,''
\emph{IEEE Trans. Med. Imag.},
vol.~40, no.~3, pp.~928--939, Mar. 2021,
doi: 10.1109/TMI.2020.3042802.

% 14
\bibitem{shang2022drac}
B.~Qian \emph{et al.},
``DRAC 2022: A public benchmark for diabetic retinopathy analysis on ultra-wide optical coherence tomography angiography images,''
\emph{Patterns},
vol.~5, no.~3, Art.~no.~100929, Mar. 2024,
doi: 10.1016/j.patter.2024.100929.

% 15
\bibitem{ronneberger2015unet}
O.~Ronneberger, P.~Fischer, and T.~Brox,
``U-Net: Convolutional networks for biomedical image segmentation,''
in \emph{Medical Image Computing and Computer-Assisted Intervention--MICCAI 2015},
LNCS 9351, pp.~234--241, 2015,
doi: 10.1007/978-3-319-24574-4\_28.

% 16
\bibitem{zhou2018unetpp}
Z.~Zhou, M.~M.~R. Siddiquee, N.~Tajbakhsh, and J.~Liang,
``UNet++: A nested U-Net architecture for medical image segmentation,''
in \emph{Deep Learning in Medical Image Analysis and Multimodal Learning for Clinical Decision Support},
LNCS 11045, pp.~3--11, 2018,
doi: 10.1007/978-3-030-00889-5\_1.

% 17
\bibitem{cao2022swinunet}
H.~Cao \emph{et al.},
``Swin-Unet: Unet-like pure transformer for medical image segmentation,''
in \emph{Computer Vision--ECCV 2022 Workshops},
LNCS 13803, pp.~205--218, 2023,
doi: 10.1007/978-3-031-25066-8\_9.

% 18
\bibitem{gu2023mamba}
A.~Gu and T.~Dao,
``Mamba: Linear-time sequence modeling with selective state spaces,''
in \emph{Proc. 1st Conf. Language Modeling (COLM)}, 2024.

% 19
\bibitem{ruan2024vmunet}
J.~Ruan, J.~Li, and S.~Xiang,
``VM-UNet: Vision Mamba UNet for medical image segmentation,''
\emph{ACM Trans. Multimedia Comput. Commun. Appl.}, 2025,
doi: 10.1145/3767748.

% 20
\bibitem{alom2018r2unet}
M.~Z. Alom, M.~Hasan, C.~Yakopcic, T.~M. Taha, and V.~K. Asari,
``Recurrent residual U-Net for medical image segmentation,''
\emph{J. Med. Imag.},
vol.~6, no.~1, Art.~no.~014006, 2019,
doi: 10.1117/1.JMI.6.1.014006.

% 21
\bibitem{oktay2018attention}
O.~Oktay \emph{et al.},
``Attention U-Net: Learning where to look for the pancreas,''
2018, arXiv:1804.03999.

% 22
\bibitem{tan2023oct2former}
X.~Tan \emph{et al.},
``OCT2Former: A retinal OCT-angiography vessel segmentation transformer,''
\emph{Comput. Methods Programs Biomed.},
vol.~233, Art.~no.~107454, May 2023,
doi: 10.1016/j.cmpb.2023.107454.

% 23
\bibitem{he2023h2former}
A.~He \emph{et al.},
``H2Former: An efficient hierarchical hybrid transformer for medical image segmentation,''
\emph{IEEE Trans. Med. Imag.},
vol.~42, no.~9, pp.~2763--2775, Sep. 2023,
doi: 10.1109/TMI.2023.3264513.

% 24
\bibitem{shit2021cldice}
S.~Shit \emph{et al.},
``clDice: A novel topology-preserving loss function for tubular structure segmentation,''
in \emph{Proc. IEEE/CVF Conf. Comput. Vis. Pattern Recognit. (CVPR)},
pp.~16555--16564, 2021,
doi: 10.1109/CVPR46437.2021.01629.

% 25
\bibitem{liu2024vmamba}
Y.~Liu \emph{et al.},
``VMamba: Visual state space model,''
in \emph{Adv. Neural Inf. Process. Syst. (NeurIPS)},
vol.~37, pp.~103\,031--103\,063, 2024.

% 26
\bibitem{ma2024umamba}
J.~Ma, F.~Li, and B.~Wang,
``U-Mamba: Enhancing long-range dependency for biomedical image segmentation,''
2024, arXiv:2401.04722.

% 27
\bibitem{wang2024mambaunet}
Z.~Wang, J.-Q. Zheng, Y.~Zhang, G.~Cui, and L.~Li,
``Mamba-UNet: UNet-like pure visual Mamba for medical image segmentation,''
2024, arXiv:2402.05079.

% 28
\bibitem{wang2024lkmunet}
J.~Wang, J.~Chen, D.~Z. Chen, and J.~Wu,
``LKM-UNet: Large kernel vision Mamba UNet for medical image segmentation,''
in \emph{Medical Image Computing and Computer Assisted Intervention--MICCAI 2024},
LNCS 15008, pp.~360--370, 2024,
doi: 10.1007/978-3-031-72111-3\_34.

% 29
\bibitem{yang2024acmamba}
V.-T. Nguyen, V.-T. Pham, and T.-T. Tran,
``AC-MambaSeg: An adaptive convolution and Mamba-based architecture for enhanced skin lesion segmentation,''
in \emph{Computational Intelligence Methods for Green Technology and Sustainable Development (GTSD 2024)},
LNNS 1195, pp.~13--26, 2024,
doi: 10.1007/978-3-031-76197-3\_2.

% 30
\bibitem{xing2024segmamba}
Z.~Xing, T.~Ye, Y.~Yang, G.~Liu, and L.~Zhu,
``SegMamba: Long-range sequential modeling Mamba for 3D medical image segmentation,''
in \emph{Medical Image Computing and Computer Assisted Intervention--MICCAI 2024},
LNCS 15008, pp.~578--588, 2024,
doi: 10.1007/978-3-031-72111-3\_54.

% 31
\bibitem{huang2024localmamba}
T.~Huang, X.~Pei, S.~You, F.~Wang, C.~Qian, and C.~Xu,
``LocalMamba: Visual state space model with windowed selective scan,''
in \emph{Computer Vision--ECCV 2024 Workshops},
LNCS 15633, pp.~12--22, 2025,
doi: 10.1007/978-3-031-91979-4\_2.

% 32
\bibitem{gros2021softseg}
C.~Gros, A.~Lemay, and J.~Cohen-Adad,
``SoftSeg: Advantages of soft versus binary training for image segmentation,''
\emph{Med. Image Anal.},
vol.~71, Art.~no.~102038, Jul. 2021,
doi: 10.1016/j.media.2021.102038.

% 33
\bibitem{lourenco2022softlabels}
J.~Louren{\c{c}}o-Silva and A.~L. Oliveira,
``Using soft labels to model uncertainty in medical image segmentation,''
in \emph{Brainlesion: Glioma, Multiple Sclerosis, Stroke and Traumatic Brain Injuries},
LNCS 12963, pp.~585--596, 2022,
doi: 10.1007/978-3-031-09002-8\_52.

% 34
\bibitem{guo2017calibration}
C.~Guo, G.~Pleiss, Y.~Sun, and K.~Q. Weinberger,
``On calibration of modern neural networks,''
in \emph{Proc. 34th Int. Conf. Mach. Learn. (ICML)},
PMLR 70, pp.~1321--1330, 2017.

% 35
\bibitem{mehrtash2020confidence}
A.~Mehrtash, W.~M. Wells, C.~M. Tempany, P.~Abolmaesumi, and T.~Kapur,
``Confidence calibration and predictive uncertainty estimation for deep medical image segmentation,''
\emph{IEEE Trans. Med. Imag.},
vol.~39, no.~12, pp.~3868--3878, Dec. 2020,
doi: 10.1109/TMI.2020.3006437.

% 36
\bibitem{brier1950verification}
G.~W. Brier,
``Verification of forecasts expressed in terms of probability,''
\emph{Monthly Weather Rev.},
vol.~78, no.~1, pp.~1--3, 1950.

% 37
\bibitem{shi2024samocta}
C.~Wang, X.~Chen, H.~Ning, and S.~Li,
``SAM-OCTA: A fine-tuning strategy for applying foundation model to OCTA image segmentation tasks,''
in \emph{Proc. IEEE Int. Conf. Acoust., Speech Signal Process. (ICASSP)},
pp.~1771--1775, 2024,
doi: 10.1109/ICASSP48485.2024.10446904.

\end{thebibliography}
\end{document}